\documentclass[10pt,letterpaper,twocolumn]{article}
\usepackage[latin9]{inputenc}
\usepackage{color}
\usepackage{array}
\usepackage{verbatim}
\usepackage{textcomp}
\usepackage{multirow}
\usepackage{amsmath}
\usepackage{amssymb}
\usepackage{graphicx}
\usepackage[unicode=true,
 bookmarks=false,
 breaklinks=false,pdfborder={0 0 1},backref=section,colorlinks=false]
 {hyperref}
\hypersetup{
 pagebackref,breaklinks,colorlinks}

\makeatletter

\pdfpageheight\paperheight
\pdfpagewidth\paperwidth

\providecommand{\tabularnewline}{\\}

\usepackage{cvpr}              
\usepackage[accsupp]{axessibility} 

\usepackage[capitalize]{cleveref}
\crefname{section}{Sec.}{Secs.}
\Crefname{section}{Section}{Sections}
\Crefname{table}{Table}{Tables}
\crefname{table}{Tab.}{Tabs.}

\def\cvprPaperID{5312} 
\def\confName{CVPR}
\def\confYear{2023}

\makeatother

\begin{document}
\title{KERM: Knowledge Enhanced Reasoning for Vision-and-Language Navigation}
\author{Xiangyang Li\textsuperscript{1,2}, Zihan Wang\textsuperscript{1,2},
Jiahao Yang\textsuperscript{1,2}, Yaowei Wang\textsuperscript{3},
Shuqiang Jiang\textsuperscript{1,2,3}\\
\textsuperscript{1}Key Lab of Intelligent Information Processing
Laboratory of the Chinese Academy of Sciences (CAS),\\ Institute
of Computing Technology, Beijing, 100190, China\\ \textsuperscript{2}University
of Chinese Academy of Sciences, Beijing, 100049, China\\ \textsuperscript{3}Peng
\textsuperscript{}Cheng Laboratory, Shenzhen, 518055, China\\\tt\footnotesize {lixiangyang@ict.ac.cn, \{zihan.wang, jiahao.yang\}@vipl.ict.ac.cn, wangyw@pcl.ac.cn, sqjiang@ict.ac.cn}}
\maketitle
\begin{abstract}
Vision-and-language navigation (VLN) is the task to enable an embodied
agent to navigate to a remote location following the natural language
instruction in real scenes. Most of the previous approaches utilize
the entire features or object-centric features to represent navigable
candidates. However, these representations are not efficient enough
for an agent to perform actions to arrive the target location. As
knowledge provides crucial information which is complementary to visible
content, in this paper, we propose a Knowledge Enhanced Reasoning
Model (KERM) to leverage knowledge to improve agent navigation ability.
Specifically, we first retrieve facts (i.e., knowledge described by
language descriptions) for the navigation views based on local regions
from the constructed knowledge base. The retrieved facts range from
properties of a single object (e.g., color, shape) to relationships
between objects (e.g., action, spatial position), providing crucial
information for VLN. We further present the KERM which contains the
purification, fact-aware interaction, and instruction-guided aggregation
modules to integrate visual, history, instruction, and fact features.
The proposed KERM can automatically select and gather crucial and
relevant cues, obtaining more accurate action prediction. Experimental
results on the REVERIE, R2R, and SOON datasets demonstrate the effectiveness
of the proposed method. The source code is available at \url{https://github.com/XiangyangLi20/KERM}.

\end{abstract}

\section{Introduction}

\label{sec:intro} Vision-and-language navigation (VLN)\ \cite{VLN-2018vision,2020reverie,qi2020object,2021airbert,zhu2021soon,zhu-etal-2022-diagnosing}
is one of the most attractive embodied AI tasks, where agents should
be able to understand natural language instructions, perceive visual
content in dynamic 3D environments, and perform actions to navigate
to the target location. Most previous methods\ \cite{2018-speaker,tan2019learning,2019reinforced,hong2021vln-bert,qi2021-road-to-know}
depend on sequential models (\textit{e.g.,}\ LSTMs and Transformers)
to continuously receive visual observations and align them with the
instructions to predict actions at each step. More recently, transformer-based
architectures\ \cite{chen2021history,chen2022think-GL,qiao2022HOP}
which make use of language instructions, current observations, and
historical information have been widely used.

\begin{figure}
\noindent %
\noindent\begin{minipage}[t]{1\columnwidth}%
\begin{center}
\includegraphics[width=1\columnwidth]{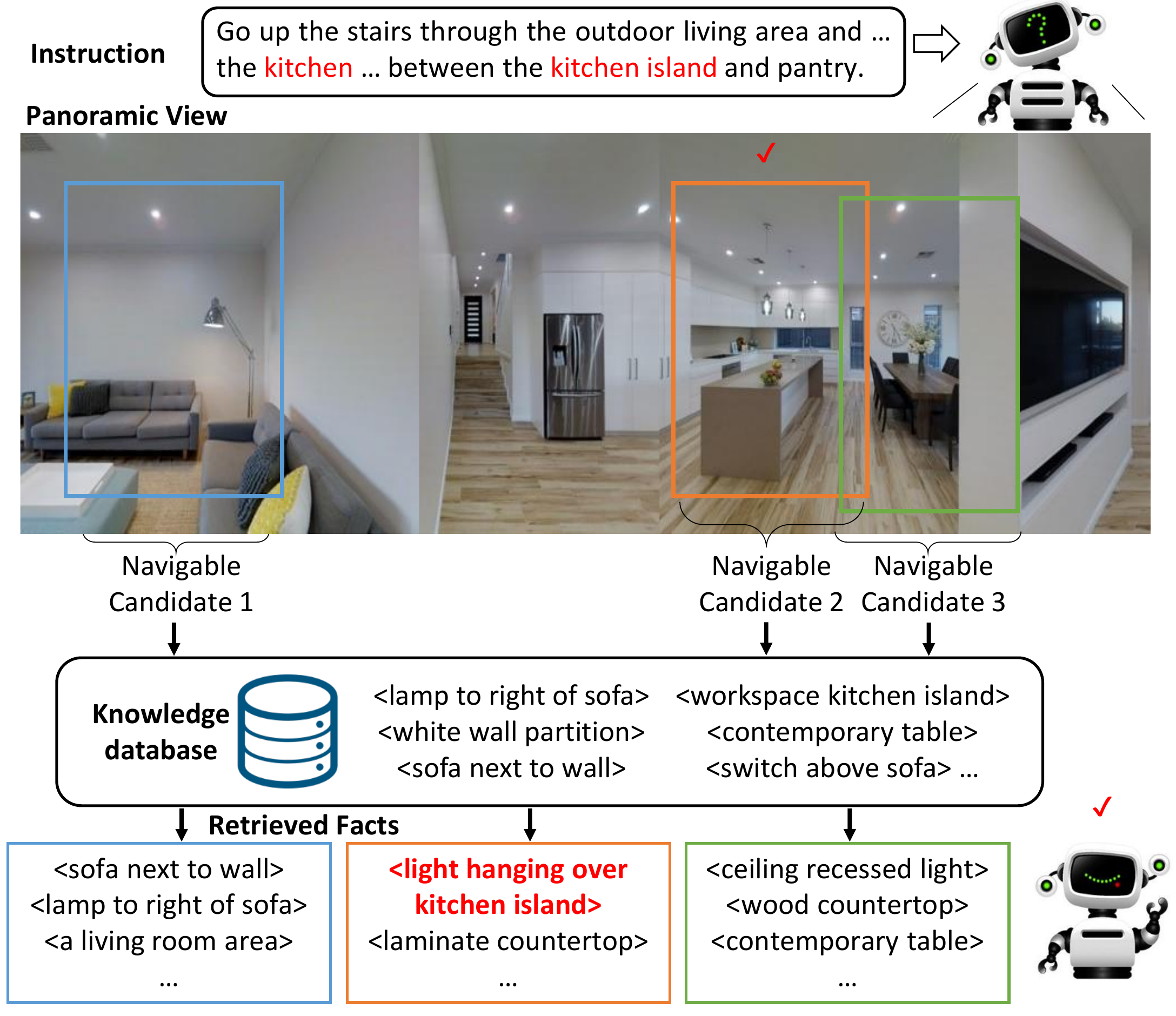} 
\par\end{center}%
\end{minipage}

\caption{Illustration of knowledge related navigable candidates, which provides
crucial information such as attributes and relationships between objects
for VLN. \label{fig:Knowledge-contains-important} Best viewed in
color.}
\end{figure}

Most of the previous approaches utilize the entire features~\cite{hao2020towards,2021airbert,chen2021history,qiao2022HOP}
or object-centric features~\cite{moudgil2021soat,gao-2021room,chen2022think-GL,an2021neighbor}
to represent navigable candidates. For example, Qi~\textit{et al.}~\cite{qi2021-road-to-know}
and Gao~\textit{et al.}~\cite{gao-2021room} encode discrete images
within each panorama with detected objects. Moudgil~\textit{et al.}~\cite{moudgil2021soat}
utilize both object-level and scene-level features to represent visual
observations. However, these representations are not efficient enough
for an agent to navigate to the target location. For example, as shown
in Figure~\ref{fig:Knowledge-contains-important}, there are three
candidates. According to the instruction and the current location,
candidate2 is the correct navigation. Based on the entire features
of a candidate view, it is hard to select the correct one, as candidate2
and candidate3 belong to the same category (\textit{i.e.,} ``dining
room''). Meanwhile, it is also hard to differentiate them from individual
objects, as ``lamp'' and ``light'' are the common components for
them.

As humans make inferences under their knowledge~\cite{gigerenzer1996reasoning},
it is important to incorporate knowledge related to navigable candidates
for VLN tasks. First, knowledge provides crucial information which
is complementary to visible content. In addition to visual information,
high-level abstraction of the objects and relationships contained
by knowledge provides essential information. Such information is indispensable
to align the visual objects in the view image with the concepts mentioned
in the instruction. As shown in Figure~\ref{fig:Knowledge-contains-important},
with the knowledge related to candidate2 (\textit{i.e.,} \textless
light hanging over kitchen island\textgreater ), the agent is able
to navigate to the target location. Second, the knowledge improves
the generalization ability of the agent. As the alignment between
the instruction and the navigable candidate is learned in limited-seen
environments, leveraging knowledge benefits the alignment in the unseen
environment, as there is no specific regularity for target object
arrangement. Third, knowledge increases the capability of VLN models.
As rich conceptual information is injected into VLN models, the correlations
among numerous concepts are learned. The learned correlations are
able to benefit visual and language alignment, especially for tasks
with high-level instructions.

In this work, we incorporate knowledge into the VLN task. To obtain
knowledge for view images, facts (\textit{i.e.,} knowledge described
by language descriptions) are retrieved from the knowledge base constructed
on the Visual Genome dataset~\cite{2017visual-genome}. The retrieved
facts by CLIP\ \cite{radford2021learning} provide rich and complementary
information for visual view images. And then, a knowledge enhanced
reasoning model (KERM) which leverages knowledge for sufficient interaction
and better alignment between vision and language information is proposed.
Especially, the proposed KERM consists of a purification module, a
fact-aware interaction module, and an instruction-guided aggregation
module. The purification model aims to extract key information in
the fact representations, the visual region representations, and the
historical representations respectively guided by the instruction.
The fact-aware interaction module allows visual and historical representations
to obtain the interaction of the facts with cross-attention encoders.
And the instruction-guided aggregation module extracts the most relevant
components of the visual and historical representations according
to the instruction for fusion.

We conduct the experiments on three VLN datasets, \textit{i.e.,} the
REVERIE~\cite{2020reverie}, SOON~\cite{zhu2021soon}, and R2R~\cite{VLN-2018vision}.
Our approach outperforms state-of-the-art methods on all splits of
these datasets under most metrics. The further experimental analysis
demonstrates the effectiveness of our method.

In summary, we make the following contributions: 
\begin{itemize}
\item We incorporate region-centric knowledge to comprehensively depict
navigation views in VLN tasks. For each navigable candidate, the retrieved
facts (\textit{i.e.,} knowledge described by language descriptions)
are complementary to visible content.
\item We propose the knowledge enhanced reasoning model (KERM) to inject
fact features into the visual representations of navigation views
for better action prediction. 
\item We conduct extensive experiments to validate the effectiveness of
our method and show that it outperforms existing methods with a better
generalization ability. 
\end{itemize}

\section{Related Work}

\begin{figure*}
\noindent %
\noindent\begin{minipage}[t]{1\columnwidth}%
\begin{center}
\includegraphics[width=0.81\paperwidth]{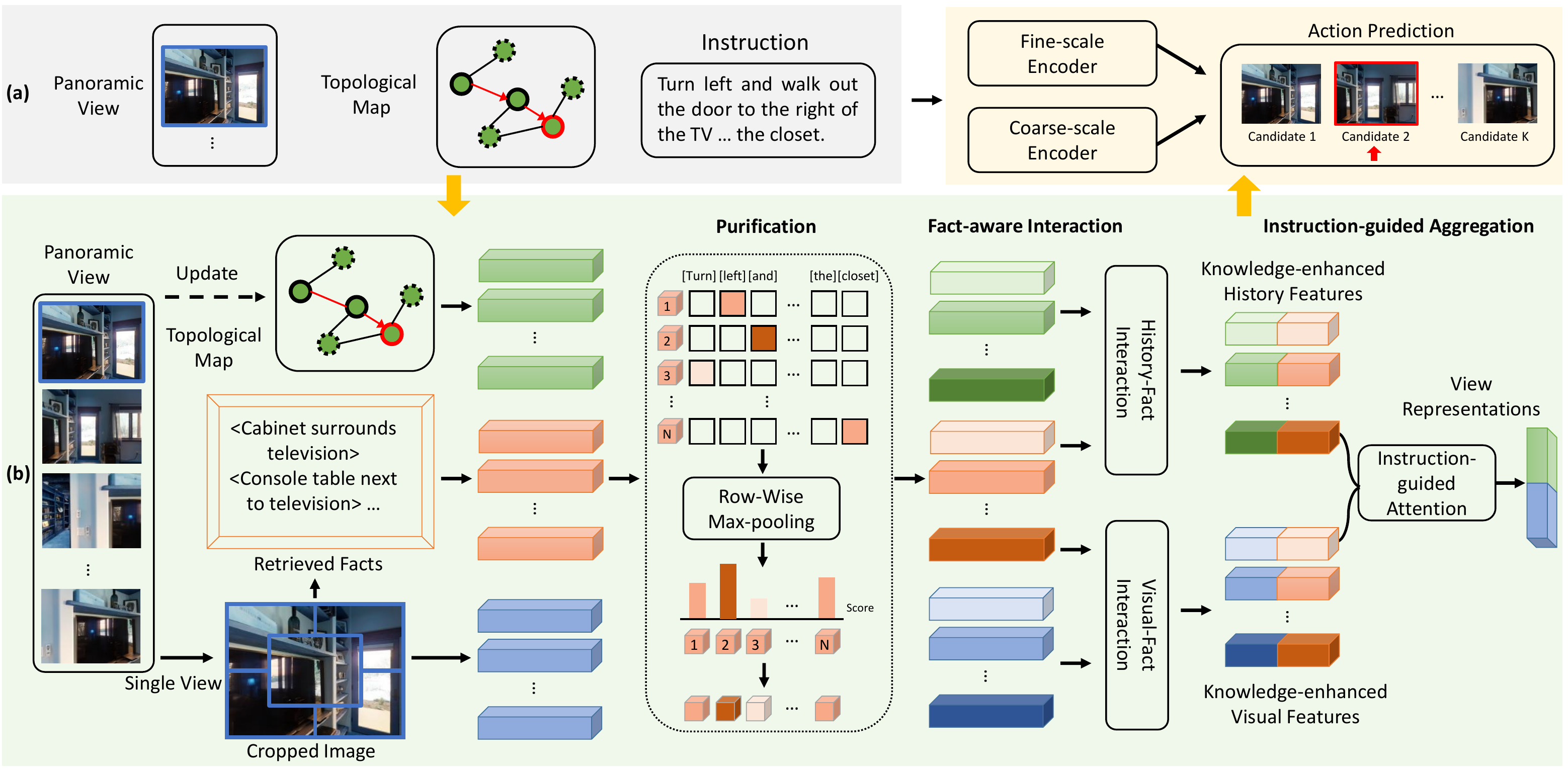}
\par\end{center}%
\end{minipage}

\caption{The overall pipeline. (a) The baseline method uses a dual-scale graph
transformer to encode the panoramic view, the topological map, and
the instruction for action prediction. (b) Our approach incorporates
retrieved facts as additional input. The knowledge enhanced representations
of each candidate view are obtained with the purification module,
the fact-aware interaction module, and the instruction-guided aggregation
module. Best viewed in color.\label{fig:The-overall-pipeline.}}
\end{figure*}

\paragraph*{Vision-and-Language Navigation.}

VLN\ \cite{VLN-2018vision,2019reinforced,tan2019learning,hong2020l-e-graph,2021structured-scene,qiao2022HOP,chen2022think-GL,Chen_2022_HM3D_AutoVLN}
has drawn significant research interests because of its practicality
for potential applications (\textit{e.g.,} healthcare robots and personal
assistants). Given natural language instructions (\textit{e.g.,}
step-by-step instructions\ \cite{VLN-2018vision,2020-RXR}, high-level
instructions\ \cite{2020reverie,zhu2021soon}, and dialog-based instructions\ \cite{dialog-navigation-2019}),
VLN tasks require the agent to navigate to the goal location. Early
methods\ \cite{2018-speaker,2019reinforced,zhu2020vision} usually
utilize recurrent neural networks (RNNs) to encode historical observations
and actions, which %
are represented as a state vector. To make further advances, numerous
approaches focus on incorporating more visual information. For example,
Hong\ \textit{et al.}\ \cite{hong2020l-e-graph}\ propose to learn
the relationships among the scene, objects, and directional clues.
In order to capture environment layouts, Wang\ \textit{et al.}\ \cite{2021structured-scene}
employ topological maps to memorize the percepts during navigation,
supporting long-term planning.

Inspired by the big success of vision-and-language pretraining\ (VLP)\ \cite{2019-lxmert,radford2021learning},
more recent approaches utilize transformer-based architectures. Hao\ \textit{et
al.}\ \cite{hao2020towards} propose PREVALENT which pretrains the
navigation model under the self-learning paradigm. To learn general
navigation-oriented textual representations, AirBERT\ \cite{2021airbert}
and HM3D-AutoVLN\ \cite{Chen_2022_HM3D_AutoVLN} create large-scale
VLN data to improve the interaction between  different modalities.
VLNBERT\ \cite{hong2021vln-bert} injects a recurrent function
into cross-modal structures to maintain time-dependent state information.
HAMT\ \cite{chen2021history}\ concatenates language instructions,
historical and current observations, and then feed them into a cross-modal
transformer to derive multi-modal action prediction. 
DUET\ \cite{chen2022think-GL} combines local observations and the
global topological map via graph transformers, facilitating action
planning and cross-modal understanding. 
In this work, we leverage mined knowledge (\textit{i.e.,} retrieved
facts) of viewpoint images to improve the generalization ability
and facilitate the alignment between vision and language for VLN.

\paragraph*{Vision-Language Reasoning with External Knowledge.}

There has been growing attention to incorporating external knowledge
for multi-modal understanding and reasoning\ \cite{qi2019ke,gao-2021room,shen2022k}.
ConceptNet\ \cite{2017conceptnet} and DBpedia\ \cite{auer2007dbpedia}
are widely used knowledge bases where concepts and relationships are
represented with nodes and edges respectively. The structured knowledge
can be represented with graph neural networks or graph transformers,
enabling interaction within visual and linguistic features. Based
on acquired knowledge, KE-GAN\ \cite{qi2019ke} utilizes the knowledge
relation loss to re-optimize parsing results. CKR\ \cite{gao-2021room}
leverages relationships among concepts according to ConceptNet to
learn correlations among rooms and object entities. Another line of
work employs unstructured knowledge which is retrieved from knowledge
bases such as Wiktionary\ \cite{meyer2012wiktionary}. For example,
K-LITE\ \cite{shen2022k} enriches natural language supervision with
knowledge descriptions, thus learning image representations that can
understand both visual concepts and their knowledge. In this work,
we leverage the knowledge that provides complementary and necessary
information for the view images to benefit VLN tasks.

\section{Method}

We address VLN in discrete environments\ \cite{VLN-2018vision,2020reverie,zhu2021soon}.
The environment is provided with a navigation connectivity graph\ $\mathcal{G},=\{\mathcal{V},\mathcal{E}\}$,
where\ $\mathcal{V}$ denotes navigable nodes and\ $\mathcal{E}$
denotes edges. Given natural language instructions, the agent needs
to explore the environment to reach the target locations.\ $\hat{L}=\{\hat{l_{i}}\}_{i=1}^{M}$
denotes the word embeddings of the instruction containing $M$ words.
At step $t$, the agent observes a panoramic view of its current node
$V_{t}$. The panoramic view contains 36 single views\ $\mathcal{S}_{t}=\{s_{i}\}_{i=1}^{36}$,
and each single view is represented by a vector $s_{i}$ accompanied
with its orientation. The navigable views of $V_{t}$ are a subset
of\ $\mathcal{S}_{t}$. For tasks with object grounding\ \cite{2020reverie,zhu2021soon},
$N$ object features\ $\mathcal{O}_{t}=\{o_{i}\}_{i=1}^{N}$\ for
a panorama are extracted with annotated bounding boxes or object detectors\ \cite{anderson2018bottom}.

\subsection{Overview of Our Approach}

The baseline follows the architecture of DUET\ \cite{chen2022think-GL},
as illustrated in Figure\ \ref{fig:The-overall-pipeline.}(a).\ A
topological map is constructed over time by adding newly observed
information to the map and updating visual representations of its
nodes. Specifically, $\mathcal{G}_{t}=\{\mathcal{V}_{t},\mathcal{E}_{t}\}$\ denotes
the topological map at step $t$, $\mathcal{G}_{t}\in\mathcal{G}$\ is
the map of the environment after $t$ navigation steps. As shown in
Figure\ \ref{fig:The-overall-pipeline.}(a), there are three kinds
of nodes in $\mathcal{V}_{t}$: visited nodes (\textit{i.e.,} nodes
with solid black border), the current node (\textit{i.e.,} the node
with solid red border), and navigable nodes (\textit{i.e.,} nodes
with dashed black border). Then at each step, all the features of
the 36 single views for the current location, the topological map,
and the instruction are fed into the dual-scale graph transformer
to predict the next location in the map or the stop action.

Our model which leverages knowledge to obtain contextualized representation
is illustrated in Figure\ \ref{fig:The-overall-pipeline.}(b). First,
facts that provide complementary information to the visual features
of each discrete view are retrieved. Then, the instruction, the topological
map, the visual view features, and the corresponding facts are fed
into the purification module, the fact-aware interaction module, and
the instruction-guided aggregation module gradually to form knowledge
enhanced representations. At last, the dual-scale graph transformer\ \cite{chen2022think-GL}
is utilized to predict the action.

\subsection{Fact Acquisition\label{subsec:Fact-Acquisition}}

To navigate to the target location, high-level abstraction of the
objects and relationships by natural language provides essential information
which is complementary to visual features. Such information is indispensable
to model the relevance between the visual objects and concepts mentioned
in instructions, facilitating the matching between candidate views
and instructions. We first construct a knowledge base and then use
the pretrained multi-modal model CLIP~ \cite{radford2021learning}
to retrieve facts for each view.

\paragraph*{Knowledge Base Construction.}

The knowledge base is the source of relevant facts that describe the
visual region. In order to obtain the rich and varied descriptions,
similar to\ \cite{kuo2022beyond}, we parse the region descriptions
from the Visual Genome\ \cite{2017visual-genome} dataset to construct
the knowledge base. Specifically, in the Visual Genome dataset, the
parsed attribute annotations take the form of ``attribute-object''
pairs and the parsed relationship annotations take the form of ``subject-predicate-object''
triplets. We convert all of the ``attribute-object'' pairs and ``subject-predicate-object''
triplets to their synset canonical forms. After removing duplicates,
we totally get 630K facts expressed by language descriptions that
are used to build our knowledge base.

\paragraph*{Fact Retrieval.}

\begin{figure}
\noindent %
\noindent\begin{minipage}[t]{1\columnwidth}%
\begin{center}
\includegraphics[width=0.96\columnwidth]{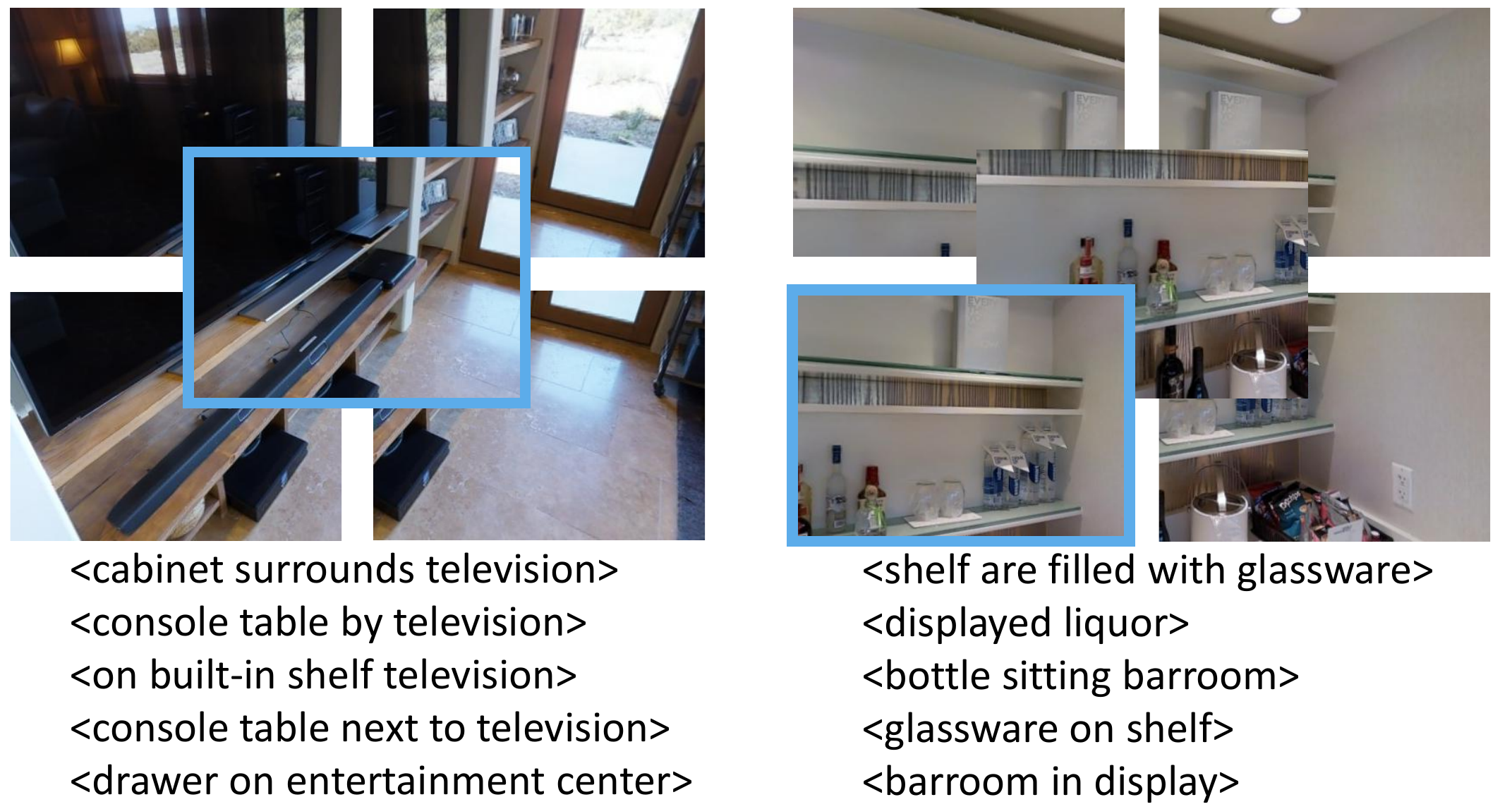} 
\par\end{center}%
\end{minipage}

\caption{Five cropped sub-regions of the view image and the retrieved top-5
facts for the sub-region in the blue box. \label{fig:Examples-of-the} }
\end{figure}

Our main goal is to obtain facts for the view images when navigating
in the environment. To this end, we crop each view image into five
sub-regions (as shown in Figure\ \ref{fig:Examples-of-the}) and
retrieve facts for these sub-regions from the knowledge base. In order
to retrieve relevant facts for the visual sub-region, we use the pretrained
model CLIP\ \cite{radford2021learning}. It consists of an image
encoder CLIP-I and a text encoder CLIP-T that encode image and text
into a joint embedding space. We use the CLIP-T to encode all facts
in the knowledge base as the search keys. The visual sub-regions are
encoded by the CLIP-I as the queries. We then search in the knowledge
base for the facts with the top-k highest cosine similarity scores.
For each sub-region, we keep five facts with the top-5 highest cosine
scores as the knowledge. Some examples of the top-5 results are shown
in Figure\ \ref{fig:Examples-of-the}.

\subsection{Knowledge Enhanced Reasoning}

At each step $t$, the visual features, the history features, the
instruction features, and the fact features are fed into our proposed
knowledge enhanced reasoning model (KERM) to support better action
prediction, as shown in Figure\ \ref{fig:The-overall-pipeline.}(b).
Specifically, for each single view $n$, the visual features $R_{n}=\{r_{i}\}_{i=1}^{5}$
are composed of 5 sub-regions where each sub-region $r_{i}$ is encoded
by CLIP-I. The fact features $E_{n}=\{e_{i}\}_{i=1}^{25}$ are the
total fact representations, where each fact $e_{i}$ is encoded by
CLIP-T. To represent historical information and the layout of the
explored environment, the history features $H_{n}$ which are composed
of all the node representations of the topological map $\mathcal{G}_{t}$\ are
also employed. As the same in\ \cite{chen2022think-GL}, word embeddings
$\hat{L}$ are fed into a multi-layer transformer to obtain the instruction
features $L$.

\subsubsection{Purification Module}

Due to the complexity of the navigation environment, a large number
of fact features within each view are not all needed by the agent
to complete the current navigation task. The agent needs more critical,
highly correlated knowledge with the current navigation instruction
to understand and characterize the environment image. Therefore, we
propose an instruction-aware fact purification module that aims to
filter out less relevant information and capture critical information
that is closely related to the current navigation task. Specifically,
we first evaluate the relevance of each fact in the view to each token
of navigation instruction by computing the fact-instruction relevance
matrix $A$ as:

\begin{equation}
A=\frac{(E_{n}W_{1})(LW_{2})^{T}}{\sqrt{d}}
\end{equation}
where $W_{1}$ and $W_{2}$ are learned parameters and $d$ is the
dimension of features. After this, we compute the row-wise max-pooling
on $A$ to evaluate the relevance of each fact to the instruction
as: 
\begin{equation}
\alpha_{n}=max\{A_{n,i}\}_{i=1}^{M}
\end{equation}
The fact features are purified by the guidance of instruction-aware
attention as: 
\begin{equation}
E_{n}^{'}=\alpha_{n}E_{n}
\end{equation}

The purified visual region features $R_{n}^{'}$ and history features
$H_{n}^{'}$ are obtained in the same way.

\subsubsection{Fact-aware Interaction Module}

After obtaining the purified features, we use a multi-layer cross-modal
transformer for vision-fact interaction. Each transformer layer consists
of a cross-attention layer and a self-attention layer. The cross-attention
between visual features and fact features is calculated as:

\begin{equation}
\tilde{R_{n}}=CrossAttn(R_{n}^{'},E_{n}^{'})
\end{equation}
And then, $\tilde{R_{n}}$ are fed into the self-attention layer to
obtain the knowledge enhanced visual features $R_{n}^{''}$. The history-fact
interaction is also conducted in this same way with another multi-layer
cross-modal transformer, forming the knowledge enhanced history features
$H_{n}^{''}$.

\subsubsection{Instruction-guided Aggregation Module}

At last, we use the instruction-guided attention to aggregate the
vision-fact features into a contextual vector $\bar{r}_{n}$ :

\begin{equation}
\eta_{n}=softmax(\frac{(R_{n}W_{3})(L_{0}W_{4})^{T}}{\sqrt{d}})
\end{equation}
\begin{equation}
\bar{r}_{n}=\sum_{i=1}^{K}\eta_{n,i}R_{n,i}^{''}
\end{equation}
where $W_{3}$ and $W_{4}$ are learned parameters and $d$ is the
dimension of features, $K$\ is the number of sub-regions (\textit{i.e.,}
$K$=5), $L_{0}$ are the feature from the {[}CLS{]} token representing
the entire instruction. With the same mechanism, we aggregate the
history-fact features into $\bar{h}_{n}$. Then we use an FFN to fuse
$\bar{r}_{n}$ and $\bar{h}_{n}$ into the knowledge enhanced representations
of view $n$, which are fed into the dual-scale graph transformer\ \cite{chen2022think-GL}
for action prediction.

\subsection{Training and Inference}

\paragraph*{Pretraining.}

As demonstrated in\ \cite{chen2021history,chen2022think-GL,qiao2022HOP},
it is beneficial to pretrain transformer-based VLN models. We utilize
the following four tasks to pretrain our KERM model.

1) Masked language modeling (MLM). We randomly mask out the words
of the instruction with a probability of 15\%, and then the masked
words $L_{m}$ are predicted with the corresponding contextual features.

2) Masked view classification (MVC). The MVC requires the model to
predict the semantic labels of masked view images. We randomly mask
out view images with a probability of 15\%. Similar to\ \cite{chen2022think-GL},
the target labels for view images are obtained by an image classification
model\ \cite{2020image-vit} pretrained on ImageNet.

3) Single-step action prediction (SAP). Given a demonstration path
$P^{*}$ and its partial $P_{<t}^{*}$, the $SAP$ loss in behavior
cloning is defined as follows:

\begin{equation}
L_{SAP}=\sum_{t=1}^{T}-log~p(a_{t}|L,P_{<t}^{*})
\end{equation}
where $a_{t}$ is the expert action of $P_{<t}^{*}$.

4) Object grounding (OG). OG is used when object annotations are available:

\begin{equation}
L_{OG}=-log~p(o^{*}|L,P_{D})
\end{equation}
where $o^{*}$ is the ground-truth object at the destination location
$P_{D}$. More details are presented in the supplementary materials.

\paragraph*{Fine-tuning and Inference.}

For fine-tuning, we use the imitation learning method\ \cite{ross2011reduction}
and the SAP loss $L_{SAP}$. Different from the pretrain process
using the demonstration path, the supervision of fine-tuning is from
a pseudo-interactive demonstrator. It selects a navigable node as
the next target node with the overall shortest distance from the current
node to the destination. For tasks with object annotations, we also
use the object grounding loss $L_{OG}$.

For inference, the model predicts an action at each step. If the action
is not the stop action, the agent performs this action and moves to
the predicted node, otherwise, the agent stops at the current node.
The agent will be forced to stop if it exceeds the maximum action
steps and then returns to the node with maximum stop probability as
its final prediction. At the predicted location, the object with the
highest prediction score is selected as the target.

\begin{table*}
\caption{Comparison with state-of-the-art methods on the REVERIE dataset.{\small{}\label{REVERIE_sota}}}

\small
\centering 
\tabcolsep=0.03cm

{\small{}\centering{}}%
\begin{tabular}{c|cccc|cc|cccc|cc|cccc|cc}
\hline 
\multirow{1}{*}{{\small{}{}Methods}} & \multicolumn{6}{c|}{{\small{}{}Val Seen}} & \multicolumn{6}{c|}{{\small{}{}Val Unseen}} & \multicolumn{6}{c}{{\small{}{}Test Unseen}}\tabularnewline
\hline 
 & \multicolumn{4}{c|}{{\small{}{}Navigation}} & \multicolumn{2}{c|}{{\small{}{}Grounding}} & \multicolumn{4}{c|}{{\small{}{}Navigation}} & \multicolumn{2}{c|}{{\small{}{}Grounding}} & \multicolumn{4}{c|}{{\small{}{}Navigation}} & \multicolumn{2}{c}{{\small{}{}Grounding}}\tabularnewline
\hline 
 & {\small{}{}TL\textdownarrow{}} & {\small{}{}OSR\textuparrow{}} & {\small{}{}SR\textuparrow{}} & {\small{}{}SPL\textuparrow{}} & {\small{}{}RGS\textuparrow{}} & {\small{}{}RGSPL\textuparrow{}} & {\small{}{}TL\textdownarrow{}} & {\small{}{}OSR\textuparrow{}} & {\small{}{}SR\textuparrow{}} & {\small{}{}SPL\textuparrow{}} & {\small{}{}RGS\textuparrow{}} & {\small{}{}RGSPL\textuparrow{}} & {\small{}{}TL\textdownarrow{}} & {\small{}{}OSR\textuparrow{}} & {\small{}{}SR\textuparrow{}} & {\small{}{}SPL\textuparrow{}} & {\small{}{}RGS\textuparrow{}} & {\small{}{}RGSPL\textuparrow{}}\tabularnewline
\hline 
{\small{}{}Seq2Seq\ \cite{VLN-2018vision}} & {\small{}{}12.88} & {\small{}{}35.70} & {\small{}{}29.59} & {\small{}{}24.01} & {\small{}{}18.97} & {\small{}{}14.96} & {\small{}{}11.07} & {\small{}{}8.07} & {\small{}{}4.20} & {\small{}{}2.84} & {\small{}{}2.16} & {\small{}{}1.63} & {\small{}{}10.89} & {\small{}{}6.88} & {\small{}{}3.99} & {\small{}{}3.09} & {\small{}{}2.00} & {\small{}{}1.58}\tabularnewline
\hline 
{\small{}{}RCM\ \cite{2019reinforced}} & {\small{}{}10.70} & {\small{}{}29.44} & {\small{}{}23.33} & {\small{}{}21.82} & {\small{}{}13.23} & {\small{}{}15.36} & {\small{}{}11.98} & {\small{}{}14.23} & {\small{}{}9.29} & {\small{}{}6.97} & {\small{}{}4.89} & {\small{}{}3.89} & {\small{}{}10.60} & {\small{}{}11.68} & {\small{}{}7.84} & {\small{}{}6.67} & {\small{}{}3.67} & {\small{}{}3.14}\tabularnewline
\hline 
{\small{}{}VLNBERT\ \cite{hong2021vln-bert}} & {\small{}{}13.44} & {\small{}{}53.90} & {\small{}{}51.79} & {\small{}{}47.96} & {\small{}{}38.23} & {\small{}{}35.61} & {\small{}{}16.78} & {\small{}{}35.02} & {\small{}{}30.67} & {\small{}{}24.90} & {\small{}{}18.77} & {\small{}{}15.27} & {\small{}{}15.68} & {\small{}{}32.91} & {\small{}{}29.61} & {\small{}{}23.99} & {\small{}{}16.50} & {\small{}{}13.51}\tabularnewline
\hline 
{\small{}{}AirBERT\ \cite{2021airbert}} & {\small{}{}15.16} & {\small{}{}49.98} & {\small{}{}47.01} & {\small{}{}42.34} & {\small{}{}32.75} & {\small{}{}30.01} & {\small{}{}18.71} & {\small{}{}34.51} & {\small{}{}27.89} & {\small{}{}21.88} & {\small{}{}18.23} & {\small{}{}14.18} & {\small{}{}17.91} & {\small{}{}34.20} & {\small{}{}30.28} & {\small{}{}23.61} & {\small{}{}16.83} & {\small{}{}13.28}\tabularnewline
\hline 
{\small{}{}HOP\ \cite{qiao2022HOP}} & {\small{}{}13.80} & {\small{}{}54.88} & {\small{}{}53.76} & {\small{}{}47.19} & {\small{}{}38.65} & {\small{}{}33.85} & {\small{}{}16.46} & {\small{}{}36.24} & {\small{}{}31.78} & {\small{}{}26.11} & {\small{}{}18.85} & {\small{}{}15.73} & {\small{}{}16.38} & {\small{}{}33.06} & {\small{}{}30.17} & {\small{}{}24.34} & {\small{}{}17.69} & {\small{}{}14.34}\tabularnewline
\hline 
{\small{}{}HAMT\ \cite{chen2021history}} & {\small{}{}12.79} & {\small{}{}47.65} & {\small{}{}43.29} & {\small{}{}40.19} & {\small{}{}27.20} & {\small{}{}15.18} & {\small{}{}14.08} & {\small{}{}36.84} & {\small{}{}32.95} & {\small{}{}30.20} & {\small{}{}18.92} & {\small{}{}17.28} & {\small{}{}13.62} & {\small{}{}33.41} & {\small{}{}30.40} & {\small{}{}26.67} & {\small{}{}14.88} & {\small{}{}13.08}\tabularnewline
\hline 
{\small{}{}DUET\ \cite{chen2022think-GL}} & {\small{}{}13.86} & {\small{}{}73.68} & {\small{}{}71.75} & {\small{}{}63.94} & {\small{}{}57.41} & {\small{}{}51.14} & {\small{}{}22.11} & {\small{}{}51.07} & {\small{}{}46.98} & {\small{}{}33.73} & {\small{}{}32.15} & {\small{}{}23.03} & {\small{}{}21.30} & {\small{}{}56.91} & {\small{}{}52.51} & {\small{}{}36.06} & {\small{}{}31.88} & {\small{}{}22.06}\tabularnewline
\hline 
\hline 
{\small{}{} KERM-pt (Ours)} & {\small{}{}14.25} & \textbf{\small{}{}74.49} & \textbf{\small{}{}71.89} & \textbf{\small{}{}64.04} & \textbf{\small{}{}57.55} & \textbf{\small{}{}51.22} & {\small{}{}22.47} & \textbf{\small{}{}53.65} & \textbf{\small{}{}49.02} & \textbf{\small{}{}34.83} & \textbf{\small{}{}33.97} & \textbf{\small{}{}24.14} & {\small{}{}18.38} & \textbf{\small{}{}57.44} & {\small{}{}52.26} & \textbf{\small{}{}37.46} & \textbf{\small{}{}32.69} & \textbf{\small{}{}23.15}\tabularnewline
\hline 
{\small{}KERM (Ours)} & {\small{}{}12.84} & \textbf{\textcolor{black}{\small{}{}79.20}} & \textbf{\textcolor{black}{\small{}{}76.88}} & \textbf{\textcolor{black}{\small{}{}70.45}} & \textbf{\textcolor{black}{\small{}{}61.00}} & \textbf{\textcolor{black}{\small{}{}56.07}} & {\small{}{}21.85} & \textbf{\textcolor{black}{\small{}{}55.21}} & \textbf{\textcolor{black}{\small{}{}50.44}} & \textbf{\textcolor{black}{\small{}{}35.38}} & \textbf{\textcolor{black}{\small{}{}34.51}} & \textbf{\textcolor{black}{\small{}{}24.45}} & {\small{}{}17.32} & \textbf{\textcolor{black}{\small{}{}57.58}} & \textbf{\textcolor{black}{\small{}{}52.43}} & \textbf{\textcolor{black}{\small{}{}39.21}} & \textbf{\textcolor{black}{\small{}{}32.39}} & \textbf{\textcolor{black}{\small{}{}23.64}}\tabularnewline
\hline 
\end{tabular}{\small\par}

{\small{} }{\small\par}
\end{table*}

\section{Experiment}

\subsection{Datasets and Evaluation Metrics}

\paragraph*{Datasets.}

We evaluate our model on the REVERIE~\cite{2020reverie}, SOON~\cite{zhu2021soon},
and R2R~\cite{VLN-2018vision}\ datasets. REVERIE contains high-level
instructions and the instructions contain 21 words on average. The
predefined object bounding boxes are provided for each panorama, and
the agent should select the correct object bounding box from candidates
at the end of the navigation path. The path length is between 4 and
7 steps. SOON also provides instructions that describe the target
locations and target objects. The average length of instructions is
47 words. However, the object bounding boxes are not provided, while
the agent needs to predict the center location of the target object.
Similar to the settings in~\cite{chen2022think-GL}, we use object
detectors~\cite{anderson2018bottom} to obtain candidate object boxes.
The path length of SOON is between 2 and 21 steps. R2R provides step-by-step
instructions and is not required to predict object locations. The
average length of instructions is 32 words and the average length
of paths is 6 steps.

\paragraph{Evaluation Metrics.}

We utilize the standard evaluation metrics~\cite{VLN-2018vision,2020reverie}
for VLN tasks to compare our method with previous approaches, including
(1) Trajectory Length (TL): the agent's average path length in meters;
(2) Navigation Error (NE): average distance in meters between the
agent's final location and the target one; (3) Success Rate (SR):
the proportion of successfully executed instructions with the NE less
than 3 meters; (4) Oracle SR (OSR): SR given the oracle stop policy;
(5) SPL: SR penalized by Path Length. For the REVERIE and SOON datasets
that require object grounding, we adopt metrics including (1) Remote
Grounding Success (RGS): proportion of successfully executed instructions;
(2) RGSPL: RGS penalized by Path Length. Except for TL and NE, all
metrics are higher the better.

\subsection{Implementation Details}

\subsubsection{Model Architectures}

We adopt the pretrained CLIP-ViT-B/16~\cite{radford2021learning}
to retrieve facts for each view. For the fact-aware interaction module,
the number of layers for the cross-modal transformer is set as 2,
with a hidden size of 768. The parameters of this module are initialized
with the pretrained LXMERT~\cite{2019-lxmert}.

We use the ViT-B/16\ \cite{2020image-vit} pretrained on ImageNet
to extract object features on the REVERIE dataset as it provides bounding
boxes. The BUTD object detector~\cite{anderson2018bottom} is utilized
on the SOON dataset to extract object bounding boxes. For the dual-scale
graph transformer~\cite{chen2022think-GL}, the number of layers
for the language encoder, panorama encoder, coarse-scale cross-modal
encoder, and fine-scale cross-modal encoder are set as 9,2,4 and 4,
respectively. And the parameters are also initialized with the pretrained
LXMERT~\cite{2019-lxmert}.

\begin{table*}
\caption{Comparison with state-of-the-art methods on the R2R dataset. $^{{\color{black}\dagger}}$\textcolor{black}{\ indicates}
reproduced results.}

\begin{centering}
\label{R2R_sota} 
\par\end{centering}
\begin{centering}
\par\end{centering}
\centering{}%
\begin{tabular}{c|cccc|cccc|cccc}
\hline 
\multirow{1}{*}{Methods} & \multicolumn{4}{c|}{Val Seen} & \multicolumn{4}{c|}{Val Unseen} & \multicolumn{4}{c}{Test Unseen}\tabularnewline
\hline 
 & TL\textdownarrow{}  & NE\textdownarrow{}  & SR\textuparrow{}  & SPL\textuparrow{}  & TL\textdownarrow{}  & NE\textdownarrow{}  & SR\textuparrow{}  & SPL\textuparrow{}  & TL\textdownarrow{}  & NE\textdownarrow{}  & SR\textuparrow{}  & SPL\textuparrow{}\tabularnewline
\hline 
Seq2Seq\ \cite{VLN-2018vision}  & 11.33  & 6.01  & 39  & -  & 8.39  & 7.81  & 22  & -  & 8.13  & 7.85  & 20  & 18\tabularnewline
\hline 
AuxRN\ \cite{zhu2020vision}  & -  & 3.33  & 70  & 67  & -  & 5.28  & 55  & 50  & -  & 5.15  & 55  & 51\tabularnewline
\hline 
PREVALENT\ \cite{hao2020towards}  & 10.32  & 3.67  & 69  & 65  & 10.19  & 4.71  & 58  & 53  & 10.51  & 5.30  & 54  & 51\tabularnewline
\hline 
EntityGraph\ \cite{hong2020l-e-graph}  & 10.13  & 3.47  & 67  & 65  & 9.99  & 4.73  & 57  & 53  & 10.29  & 4.75  & 55  & 52\tabularnewline
\hline 
VLNBERT\ \cite{hong2021vln-bert}  & 11.13  & 2.90  & 72  & 68  & 12.01  & 3.93  & 63  & 57  & 12.35  & 4.09  & 63  & 57\tabularnewline
\hline 
AirBERT\ \cite{2021airbert}  & 11.09  & 2.68  & 75  & 70  & 11.78  & 4.01  & 62  & 56  & 12.41  & 4.13  & 62  & 67\tabularnewline
\hline 
HOP\ \cite{qiao2022HOP}  & 11.26  & 2.72  & 75  & 70  & 12.27  & 3.80  & 64  & 57  & 12.68  & 3.83  & 64  & 59\tabularnewline
\hline 
HAMT\ \cite{chen2021history}  & -  & -  & -  & -  & 11.46  & 2.29  & 66  & 61  & 12.27  & 3.93  & 65  & 60\tabularnewline
\hline 
DUET\ \cite{chen2022think-GL}  & -  & -  & -  & -  & 13.94  & 3.31  & 72  & 60  & 14.73  & 3.65  & 69  & 59\tabularnewline
\hline 
\hline 
DUET\textcolor{black}{${\color{red}^{{\color{black}\dagger}}}$} & 12.30  & 2.28  & 78.84  & 72.89  & 13.94  & 3.31  & 71.52  & 60.42  & 14.74  & 3.65  & 69.25  & 58.68\tabularnewline
\hline 
KERM (Ours)  & 12.16  & \textbf{2.19}  & \textbf{79.73}  & \textbf{73.79}  & 13.54  & \textbf{3.22}  & \textbf{71.95}  & \textbf{60.91}  & 14.60  & \textbf{3.61}  & \textbf{69.73}  & \textbf{59.25}\tabularnewline
\hline 
\end{tabular}
\end{table*}

\subsubsection{Training Details}

For pretraining, we set the batch size as 16 using 4 NVIDIA RTX3090
GPUs. For the REVERIE dataset, we combine the original dataset with
augmented data synthesized by DUET~\cite{chen2022think-GL} to pretrain
our model with 100k iterations. Then we fine-tune the pretrained model
with the batch size of 2 for 20k iterations on 4 NVIDIA RTX3090 GPUs.
For the SOON dataset, we only use the original data with automatically
cleaned object bounding boxes, sharing the same settings in DUET~\cite{chen2022think-GL}.
We pretrain the model with 40k iterations. And then we fine-tune the
pretrained model with the batch size of 1 for 20k iterations on 4
NVIDIA RTX3090 GPUs. For the R2R dataset, additional augmented R2R
data in\ \cite{hao2020towards} is used in pretraining. We pretrain
our model with 200k iterations. Then we fine-tune the pretrained model
with the batch size of 2 for 20k iterations on 4 NVIDIA RTX3090 GPUs.
For all the datasets, the best epoch is selected by SPL on the val
unseen split.

\subsection{Comparison to State-of-the-Art Methods}

\noindent Table~\ref{REVERIE_sota},\ref{R2R_sota},\ref{SOON_sota}
 illustrate the results of our method on the REVERIE, R2R and SOON
 datasets. Our approach achieves state-of-the-art performance under
most metrics on all the seen/unseen splits of all three datasets,
demonstrating the effectiveness of the proposed model. In particular,
as shown in Table~\ref{REVERIE_sota}, our model outperforms the
previous DUET~\cite{chen2022think-GL} by 5.52\% on SR, 6.51\% on
SPL, and 4.93\% on RGSPL for val seen split of the REVERIE dataset.
As shown in Table~\ref{R2R_sota} and \ref{SOON_sota}, our approach
also shows performance gains on the R2R and SOON datasets, but not
as pronounced on the REVERIE dataset. It is mainly because of the
smaller size of data and more complex instructions, which make the
generalization ability of our knowledge enhancement method not fully
exploited.

In addition, as shown in Table~\ref{REVERIE_sota}, we show the results
of our model with two training strategies. KERM denotes the full pipeline
which is composed of pretraining and fine-tuning. For KERM, our model
is first pretrained with augmented data, where the parameters are
initialized with LXMERT\ \cite{2019-lxmert}. And then, it is fine-tuned
with standard data. KERM-pt denotes the pipeline which only fine-tunes
our knowledge enhanced reasoning model with standard data, based on
the baseline method (\textit{i.e.,}\ the pretrained model in DUET\cite{chen2022think-GL}).
Both models outperform the previous methods, further illustrating
the effectiveness of the proposed knowledge enhancement method. Then,
with the help of pretraining, KERM achieves better performance, reflecting
that the generalization ability of our method can be further strengthened
by pretraining with larger data.

\begin{table}
\caption{Comparison with state-of-the-art methods on the val unseen split of
the SOON dataset.}
\tabcolsep=0.08cm

\centering{}%
\begin{tabular}{c|ccccc}
\hline 
Method  & TL\textdownarrow{}  & OSR\textuparrow{}  & SR\textuparrow{}  & SPL\textuparrow{}  & RGSPL\textuparrow{}\tabularnewline
\hline 
GBE\ \cite{zhu2021soon}  & 28.96  & 28.54  & 19.52  & 13.34  & 1.16\tabularnewline
\hline 
DUET\ \cite{chen2022think-GL}  & 36.20  & 50.91  & 36.28  & 22.58  & 3.75\tabularnewline
\hline 
\hline 
KERM (Ours)  & 35.83  & \textbf{51.62}  & \textbf{38.05}  & \textbf{23.16}  & \textbf{4.04}\tabularnewline
\hline 
\end{tabular}\label{SOON_sota} 
\end{table}

\subsection{Further Analysis}

\paragraph*{Ablation Study.}

As shown in Table~\ref{ablation_study}, we evaluate the impact
of the key components of our model on the val unseen split of REVERIE.
First, the results in row 2-4 are better than row 1, which demonstrate
that all these modules are beneficial for navigation. Then, the results
in row 5 are better than row 2,3, validating the effect of the knowledge
enhanced representations in the vision-fact interaction or the history-fact
interaction. Finally, the results in row 5 are better than row 4,
validating the effectiveness of purification operations. With the
purification, interaction, and aggregation modules, our method obtains
the best performance.
\begin{center}
\begin{table}
\caption{Ablation study results on val unseen split of the REVERIE dataset.
\textquotedblleft Pur.\textquotedblright{} denotes the purification
module. \textquotedblleft VF-Int.\textquotedblright{} and \textquotedblleft HF-Int.\textquotedblright{}
denote the vision-fact and history-fact interaction module respectively.}
\noindent\begin{minipage}[t]{1\columnwidth}%
\tabcolsep=0.06cm 
\begin{center}
\begin{tabular}{ccc|ccccc}
\hline 
Pur.  & VF-Int.  & HF-Int.  & OSR\textuparrow{}  & SR\textuparrow{}  & SPL\textuparrow{}  & RGS\textuparrow{}  & RGSPL\textuparrow{}\tabularnewline
\hline 
 &  &  & 51.07  & 46.98  & 33.73  & 32.15  & 23.03\tabularnewline
\hline 
$\checkmark$  &  & $\checkmark$  & 53.78  & 49.33  & 35.01  & 33.80  & 23.89\tabularnewline
\hline 
$\checkmark$  & $\checkmark$  &  & 53.53  & 48.96  & 34.70  & 33.63  & 23.58\tabularnewline
\hline 
 & $\checkmark$  & $\checkmark$  & 54.01  & 49.43  & 34.79  & 33.22  & 23.80\tabularnewline
\hline 
$\checkmark$  & $\checkmark$  & $\checkmark$  & \textbf{\textcolor{black}{55.21}}  & \textbf{\textcolor{black}{50.44}}  & \textbf{\textcolor{black}{35.38}}  & \textbf{\textcolor{black}{34.51}}  & \textbf{\textcolor{black}{24.45}}\tabularnewline
\hline 
\end{tabular}
\par\end{center}%
\end{minipage}\label{ablation_study} 
\end{table}
\par\end{center}

\begin{figure*}
\noindent %
\noindent\begin{minipage}[t]{1\columnwidth}%
\begin{center}
\includegraphics[width=0.81\paperwidth]{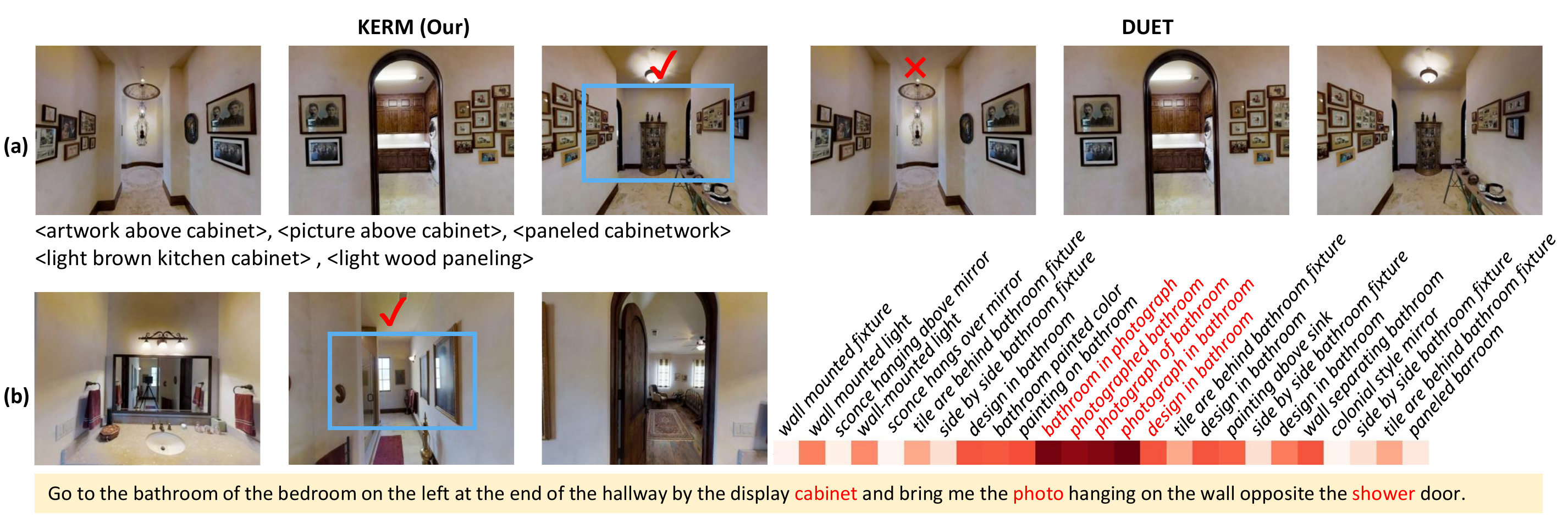} 
\par\end{center}%
\end{minipage}

\caption{Visualization of navigation examples. The sentence within the yellow
box is the navigation instruction for the agent. (a) shows a comparison
where our KERM chooses the right location while the baseline model
makes the wrong choice. (b) illustrates the weights for the 25 facts
after the fact purification when the agent navigates to the right
location. The facts for the sub-region within the blue box are annotated
in red. The results show that our model can automatically select the
relevant facts.\label{fig:Visualisation}}
\end{figure*}

\paragraph*{Fact vs. Object.}

In order to illustrate that region-based knowledge is superior to
the object-centric information, we compare our method with the model
utilizing object enhanced representations. Specifically, for each
cropped region image, we use VinVL\ \cite{zhang2021vinvl} to detect
objects and take five objects with the highest scores as substitutes
for facts. For fair comparison with retrieved facts, we prompt each
object label by adding ``an image of a'' before the label to form
a short sentence and then use CLIP-T to obtain object-centric features.
The purification module and fact-aware interaction module are the
same as the fact-based approach. As shown in Table~\ref{factvsobj},
we compare the baseline method and the methods of adding object features
and fact features on the val unseen split of REVERIE. We can see that
the object features are beneficial to navigation, but they are still
inferior to fact representations. This is consistent with our assumption
that object tags are not sufficient to provide strong generalization
performance compared to facts which have a much larger semantic feature
space.

\begin{table}
\caption{Comparison between fact and object.}
\tabcolsep=0.08cm%
\noindent\begin{minipage}[t]{1\columnwidth}%
\begin{center}
\begin{tabular}{c|ccccc}
\hline 
 & OSR\textuparrow{}  & SR\textuparrow{}  & SPL\textuparrow{}  & RGS\textuparrow{}  & RGSPL\textuparrow{}\tabularnewline
\hline 
Baseline  & 51.07  & 46.98  & 33.73  & 32.15  & 23.03\tabularnewline
\hline 
+ Object  & 54.27  & 49.73  & 33.95  & 33.91  & 23.40\tabularnewline
\hline 
+ Fact (KERM)  & \textbf{\textcolor{black}{55.21}}  & \textbf{\textcolor{black}{50.44}}  & \textbf{\textcolor{black}{35.38}}  & \textbf{\textcolor{black}{34.51}}  & \textbf{\textcolor{black}{24.45}}\tabularnewline
\hline 
\end{tabular}
\par\end{center}%
\end{minipage}\label{factvsobj} 
\end{table}

\paragraph*{Number of Regions.}

As shown in Table~\ref{number_region}, we explore the impact of
the number of cropped sub-regions on navigation performance on the
val unseen split of REVERIE. It is demonstrated that the strategy
of five sub-regions yields the best results. It is mainly because
fewer but larger sub-regions are difficult to retrieve more fine-grained
facts, while more but smaller regions are too fragmented to contain
all the complete parts that can retrieve accurate facts.

\begin{table}
\caption{The effect of the number of cropped sub-regions.}
\noindent\begin{minipage}[t]{1\columnwidth}%
\begin{center}
\begin{tabular}{c|ccccc}
\hline 
Number  & OSR\textuparrow{}  & SR\textuparrow{}  & SPL\textuparrow{}  & RGS\textuparrow{}  & RGSPL\textuparrow{}\tabularnewline
\hline 
1  & 55.69  & 49.67  & 34.46  & 34.05  & 23.89\tabularnewline
\hline 
5  & \textbf{\textcolor{black}{55.21}}  & \textbf{\textcolor{black}{50.44}}  & \textbf{\textcolor{black}{35.38}}  & \textbf{\textcolor{black}{34.51}}  & \textbf{\textcolor{black}{24.45}}\tabularnewline
\hline 
9  & 53.59  & 48.99  & 34.87  & 33.14  & 23.79\tabularnewline
\hline 
\end{tabular}
\par\end{center}%
\end{minipage}\label{number_region} 
\end{table}

\subsection{Qualitative Results}

Examples in the val unseen split of REVERIE are illustrated in Figure~\ref{fig:Visualisation}.
The sentence within the yellow box is the navigation instruction of
these examples and the red color words are the landmarks (\textit{e.g.,}
``cabinet" and ``shower door") or target objects (\textit{e.g.,}
``photo"). Figure~\ref{fig:Visualisation}(a) shows an example
where our KERM chooses the correct navigation direction by capturing
the facts associated with the ``cabinet", as shown in the left bottom
of Figure~\ref{fig:Visualisation}(a). The baseline model makes the
wrong choice without the facts. The left of Figure~\ref{fig:Visualisation}(b)
shows the cropped region in the blue box where target objects (\textit{i.e.,}
objects belong to ``photo") are contained. In the right of Figure~\ref{fig:Visualisation}(b),
a heat map is utilized to visualize the weight distribution after
the purification process of the 25 facts mined from the target view
image. Particularly, we use red color to highlight the five facts
only mined from the region in the blue box. We can see that the facts
annotated in red contain the words related to the target object (\textit{i.e.,}
``photo hanging on the wall opposite the shower door") in the navigation
instruction, such as ``photograph" and ``bathroom". It is worth
noting that these facts related to the target object have almost the
highest purification weights, which demonstrates that our model can
automatically select the relevant facts, thus obtain better performance.

\section{Conclusion}

In this paper, we propose KERM, a knowledge enhanced reasoning model
to improve the generalization ability and performance for VLN. Our
work utilizes a large number of external fact descriptions to build
a knowledge base and introduces the CLIP model to retrieve the facts
(\textit{i.e.,} knowledge described by language descriptions) for
view images. We further design a knowledge enhanced reasoning approach
with the process of purification, interaction, and aggregation that
automatically obtains contextual information for action prediction.
We illustrate the good interpretability of KERM and provide case study
in deep insights. Our approach achieves excellent improvement on
many VLN tasks, demonstrating that leveraging knowledge is a promising
direction in improving VLN and Embodied AI. For future work, we will
improve our KERM with larger training data and employ it on VLN in
continuous environments.

\paragraph*{Acknowledgment. }

This work was supported in part by the National Natural Science Foundation
of China under Grants 62125207, 62102400, 62272436, U1936203, and
U20B2052, in part by the National Postdoctoral Program for Innovative
Talents under Grant BX20200338, and in part by the Peng Cheng Laboratory
Research Project under Grant PCL2021A07.

{\small{}\bibliographystyle{ieee_fullname}
\bibliography{vln23cvpr}
}{\small\par}

\appendix

\section*{Appendix}

Section\ \ref{sec:A.-Model-Details} provides additional details
for the pretraining of our overall framework. The experimental results
for the improvement based on the fine-scale encoder and the coarse-scale
encoder are illustrated in Section\ \ref{sec:Improvements-On-Different}.
Section\ \ref{sec:More-Qualitative-Results} presents more qualitative
examples.

\section{Pretraining Details\label{sec:A.-Model-Details}}

To pretrain our proposed KERM, we utilize four auxiliary tasks. Besides
the behavior cloning tasks, \textit{i.e., }single-step action prediction
(SAP) and object grounding (OG), the masked language modeling (MLM)
and masked region classification (MRC) are utilized. In the following,
these two tasks are described.
\begin{itemize}
\item Masked language modeling (MLM). MLM aims to learn language representations
by masking parts of the text and predicting them with the contextual
information. The inputs of this task are pairs of language instruction
$L$\ and the corresponding demonstration path $P$. As our method
utilizes the dual-scale graph transformer\ \cite{chen2022think-GL}
for action prediction, we also average the embeddings of the fine-scale
and coarse-scale encoders and then a network with two fully-connected
layers is used to predict the target word. Similar to previous approaches\ \cite{hao2020towards,qiao2022HOP},
we randomly mask out the instruction words with a probability of 15\%.
This task is optimized by minimizing the negative log-likelihood of
the original words:
\[
L_{MLM}=-log(w_{i}|L_{m},P)
\]
where $L_{m}$ is the masked instruction and $w_{i}$\ is the label
of the masked word.
\item Masked region classification (MRC). MRC requires the model to predict
the semantic labels of masked view images according to the instruction,
unmasked view images, and the corresponding features in the topological
map. With the same settings in DUET\ \cite{chen2022think-GL}, we
randomly mask out view images and objects in the last observation
of the corresponding demonstration path $P$ with a probability of
15\% in the fine-scale encoder. The visual features for the masked
images or objects are set to zero, while their position embeddings
are preserved. The target semantic labels for view images are predicted
by an image classification model\ \cite{2020image-vit} pretrained
on ImageNet, and the labels for the objects are obtained by an object
detector\ \cite{anderson2018bottom} pretrained on the Visual Genome
dataset\ \cite{2017visual-genome}. Similar to\ \cite{chen2022think-GL},
we use a two-layer fully-connected network to predict the semantic
labels of masked visual tokens, and the KL divergence between the
predicted and target probability distribution of each mask token is
minimized.
\end{itemize}
\begin{figure*}
\noindent\begin{minipage}[t]{1\columnwidth}%
\begin{center}
\includegraphics[width=0.82\paperwidth]{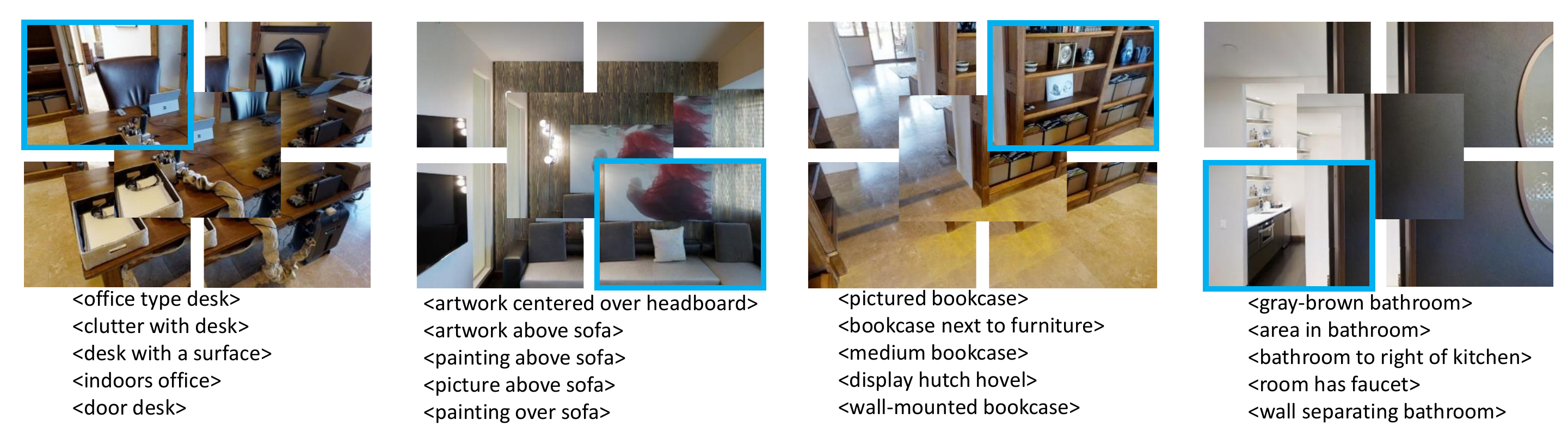}
\par\end{center}%
\end{minipage}

\caption{Examples of the retrieved facts for view images. Each view image is
cropped into five sub-regions. We show the top-5 facts for the sub-regions
in the blue box.\label{fig:Examples-of-the-1}}
\end{figure*}

\section{Improvements on Different Scales\label{sec:Improvements-On-Different}}

We also evaluate our proposed method on the settings with only the
fine-scale encoder and the coarse-scale encoder separately, as illustrated
in Table\ \ref{tab:The-results-of}. When only using the fine-scale
encoder to predict the action, our KERM-fine significantly outperforms
DUET-fine. For example, the Success Rate (SR) is improved from 28.86\%
to 30.80\%. The trend on the coarse-scale encoder is also the same.
With both the fine-scale and the coarse-scale encoders, our KERM improves
the SR by 3.3\%. The results demonstrate the effectiveness of our
method.

\begin{table}
\caption{The results of different scales and dual-scale fusion on the val unseen
split of the REVERIE dataset. \label{tab:The-results-of}}

\tabcolsep=0.08cm

\noindent\begin{minipage}[t]{1\columnwidth}%
\begin{center}
\begin{tabular}{c|ccccc}
\hline 
 & OSR\textuparrow{} & SR\textuparrow{} & SPL\textuparrow{} & RGS\textuparrow{} & RGSPL\textuparrow{}\tabularnewline
\hline 
DUET-fine & 30.96 & 28.86 & 23.57 & 20.39 & 16.64\tabularnewline
\hline 
KERM-fine & 34.40 & 30.80 & 24.83 & 21.68 & 17.56\tabularnewline
\hline 
DUET-coarse & 46.44 & 36.52 & 25.98 & - & -\tabularnewline
\hline 
KERM-coarse & 46.38 & 37.38 & 26.32 & - & -\tabularnewline
\hline 
DUET & 51.07 & 46.98 & 33.73 & 32.15 & 23.03\tabularnewline
\hline 
KERM & 55.21 & 50.44 & 35.38 & 34.51 & 24.45\tabularnewline
\hline 
\end{tabular}
\par\end{center}%
\end{minipage}
\end{table}

Furthermore, we investigate the effect of the strategy that the agent
selects the most likely viewpoint in the navigation history if the
timesteps go above a certain threshold. For the fair comparison
with previous approaches, we apply the same settings as\ \cite{chen2022think-GL}.
Specifically,  we set the maximum action steps as 15 for REVERIE
and R2R, and\textbf{ }20 for SOON. Table\ \ref{tab:The-proportion-of}
illustrates the proportion of the episodes  terminated with the maximum
action steps. For example, on the val unseen split of REVERIE, \textquotedblleft 560/3521
(15.90\%)\textquotedblright{} represents that this split has 3521
episodes, and 560 of them are terminated by the  criterion of the
termination policy (TP), making up a proportion of 15.90\%.  Moreover,
Table\ \ref{tab:The-influence-of-TP} shows the results of our KERM
and the policy that the agent stops at the last visited location
(\textit{i.e.}, KERM w/o TP). The results illustrate that the employed
TP has small influence on REVERIE and R2R, while has great influence
on SOON. This is because that the average hop of the trajectories
on SOON is longer with more complex language instructions.

\begin{table}

\caption{Statistics of episodes with the maximum action steps.\label{tab:The-proportion-of}}

\small
\tabcolsep=0.04cm
\centering

\centering{}%
\begin{tabular}{c|c|c|c}
\hline 
 & Val Seen & Val Unseen & Test\tabularnewline
\hline 
REVERIE & 31/1423 (2.18\%) & 560/3521 (15.90\%) & 898/6292 (14.27\%)\tabularnewline
\hline 
R2R & 10/1021 (0.98\%) & 51/2349 (2.17\%) & 132/4173 (3.16\%)\tabularnewline
\hline 
SOON & - & 888/2261 (39.27\%) & 1423/3080 (46.20\%)\tabularnewline
\hline 
\end{tabular}
\end{table}

\begin{table}

\caption{Influence of the termination policy on the val unseen split.\label{tab:The-influence-of-TP}}

\small
\tabcolsep=0.04cm
\centering
\begin{centering}
\begin{tabular}{c|c|c|c|c|c}
\hline 
 &  & OSR & SR & SPL & RGSPL\tabularnewline
\hline 
\multirow{2}{*}{REVERIE} & KERM & \textbf{55.21} & \textbf{50.44} & \textbf{35.38} & \textbf{24.45}\tabularnewline
\cline{2-6} \cline{3-6} \cline{4-6} \cline{5-6} \cline{6-6} 
 & KERM w/o TP & 55.21 & 49.96 & 35.29 & 24.25\tabularnewline
\hline 
\multirow{2}{*}{R2R} & KERM & \textbf{80.42} & \textbf{71.95} & \textbf{60.91} & -\tabularnewline
\cline{2-6} \cline{3-6} \cline{4-6} \cline{5-6} \cline{6-6} 
 & KERM w/o TP & 80.42 & 71.90 & 60.77 & -\tabularnewline
\hline 
\multirow{2}{*}{SOON} & KERM & \textbf{51.62} & \textbf{38.05} & \textbf{23.16} & \textbf{4.04}\tabularnewline
\cline{2-6} \cline{3-6} \cline{4-6} \cline{5-6} \cline{6-6} 
 & KERM w/o TP & 51.62 & 35.23 & 21.53 & 3.52\tabularnewline
\hline 
\end{tabular}
\par\end{centering}

\end{table}

\begin{figure}
\noindent\begin{minipage}[t]{1\columnwidth}%
\begin{center}
\includegraphics[width=1\columnwidth]{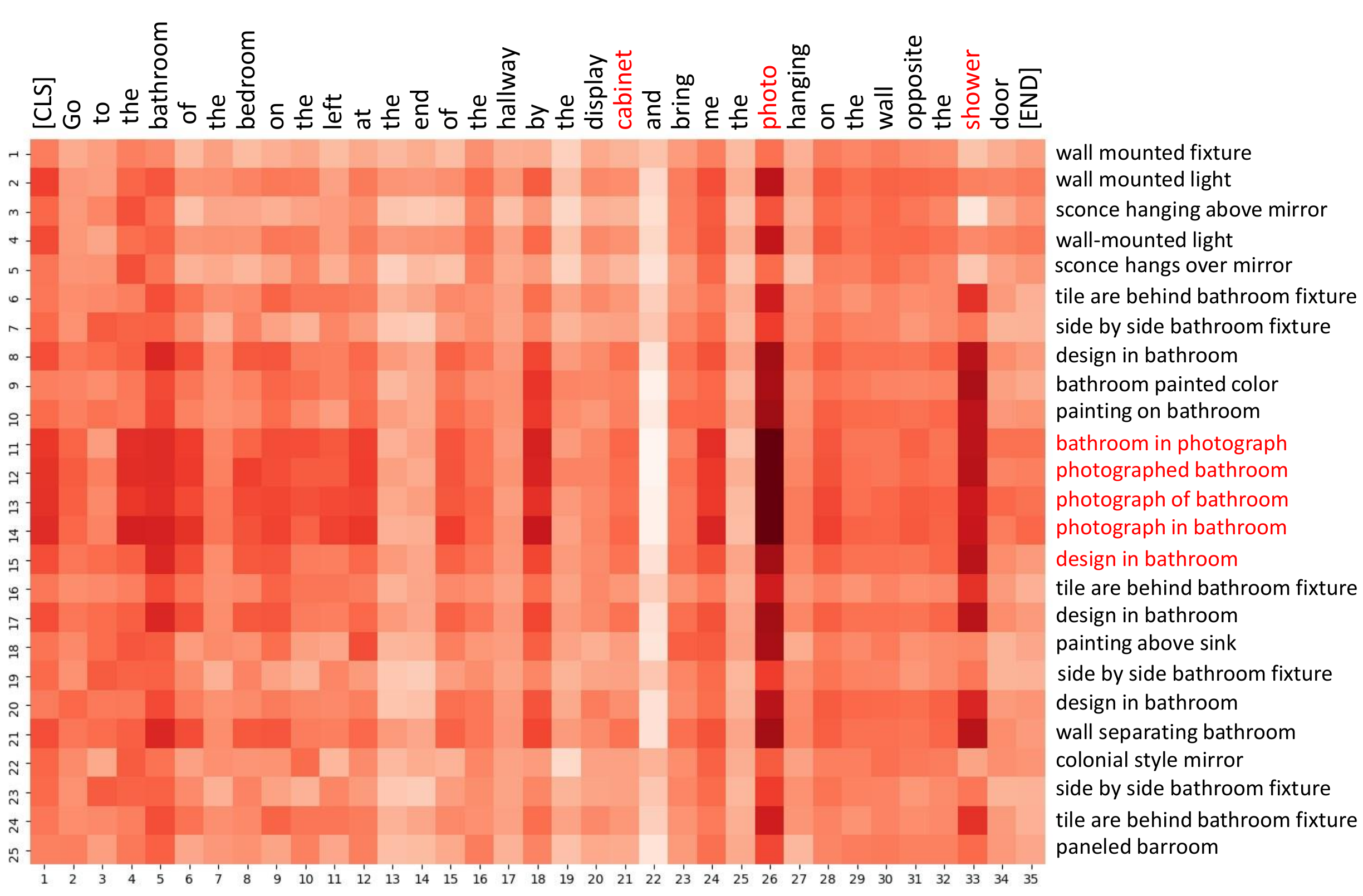}
\par\end{center}%
\end{minipage}

\caption{Illustration of the weights for the 25 facts during fact purification.
Best viewed in color.\label{fig:Illustration-of-the}}
\end{figure}

\section{More Qualitative Results\label{sec:More-Qualitative-Results}}

Figure\ \ref{fig:Examples-of-the-1} illustrates the retrieved facts
for view images. The retrieved facts provide crucial information (\textit{e.g.,}
attributes and relationships between objects) which are complementary
to visual features. Figure\ \ref{fig:Illustration-of-the} demonstrates
the weights for each fact corresponding to each word in the instruction
during fact purification. It is illustrated that our model can automatically
select the relevant facts to make better action prediction.

\end{document}